\documentclass{article}

     \PassOptionsToPackage{numbers, compress}{natbib} 

     \usepackage[final]{tackling_climate_workshop_style}

\usepackage[utf8]{inputenc} 
\usepackage[T1]{fontenc}    
\usepackage{hyperref}       
\usepackage{url}            
\usepackage{booktabs}       
\usepackage{amsfonts}       
\usepackage{nicefrac}       
\usepackage{microtype}      
\usepackage{graphicx}       
\usepackage{subcaption}     
\usepackage{float}

\title{Detection and Simulation of Urban Heat Islands Using a Fine-Tuned Geospatial Foundation Model for Microclimate Impact Prediction}

\author{%
  Jannis Fleckenstein\thanks{Equal contribution.} \\
  IBM Research – Europe \\
  \texttt{Jannis.Fleckenstein@ibm.com}
  \And
  David Kreismann\footnotemark[1] \\
  IBM Research – Europe \\
  \texttt{David.Kreismann@ibm.com}
  \And
  Tamara Rosemary Govindasamy \\
  IBM Research – Africa \\
  \texttt{tamara.govindasamy@ibm.com}
  \And
  Thomas Brunschwiler \\
  IBM Research – Europe \\
  \texttt{tbr@zurich.ibm.com}
  \And
  Etienne Vos \\
  IBM Research – Africa \\
  \texttt{etienne.vos@ibm.com}
  \And
  Mattia Rigotti \\
  IBM Research – Europe \\
  \texttt{mrg@zurich.ibm.com}
}

\begin{document}

\maketitle

\begin{abstract}
As urbanization and climate change progress, urban heat island effects are becoming more frequent and severe. To formulate effective mitigation plans, cities require detailed air temperature data, yet conventional machine learning models with limited data often produce inaccurate predictions, particularly in underserved areas. Geospatial foundation models trained on global unstructured data offer a promising alternative by demonstrating strong generalization and requiring only minimal fine-tuning. In this study, an empirical ground truth of urban heat patterns is established by quantifying cooling effects from green spaces and benchmarking them against model predictions to evaluate the model’s accuracy. The foundation model is subsequently fine-tuned to predict land surface temperatures under future climate scenarios, and its practical value is demonstrated through a simulated in-painting that highlights its role for mitigation support. The results indicate that foundation models offer a powerful way for evaluating urban heat island mitigation strategies in data-scarce regions to support more climate-resilient cities.
\end{abstract}

\section{Introduction}
As cities grow, increasing building and population densities drive land use changes that
impact the local energy balance and shape the microclimate conditions of
urban areas \cite{NURWANDA2020101772}. With over 70\% of the global population projected to live in cities by
2050, the urban heat island (UHI) effect has become a major concern \cite{LIN2023109910, su14159234}. Factors
such as heat-retaining materials, vegetation loss and dense structures can
cause cities to be over 5~°C warmer than their surrounding areas \cite{book}. These temperature
increases lead to greater energy consumption, more heat-related health issues, and declining
air quality \cite{SANTAMOURIS2015119, tanUrbanHeatIsland2010}.\\
One major challenge in addressing UHI impacts is the lack of high-resolution, timely air
temperature data, which is essential for early warnings and effective heat risk mitigation
plans \cite{caoModelingIntraurbanDifferences2022, kousisIntraurbanMicroclimateInvestigation2021}. 
Current forecasting approaches often require large amounts of input
data, specialized expertise, and significant computational resources, making real-time
forecasting at fine scales largely inaccessible \cite{deridderUrbClimFastUrban2015, powersWeatherResearchForecasting2017}. 
Data limitations from sparse stations and satellite revisit intervals further constrain accuracy \cite{varentsovMachineLearningSimulation2023, zhaoOverviewApplicationsEarth2022}. 
While artificial intelligence (AI)-based models offer increased flexibility, they are often limited by
their dependence on large labeled datasets and poor generalization across domains \cite{10.1007/978-3-030-71704-9_65}.

In response to these limitations, recent developments in geospatial foundation models (GFMs), show strong generalization across spatial resolutions and regions with minimal fine-tuning, making them a promising alternative for urban climate analysis \cite{mai2023opportunitieschallengesfoundationmodels}.
Building on this potential, this study explores the use of a fine-tuned GFM by establishing an empirical ground truth of how urban green spaces affect temperature through internal and spillover cooling in UHI areas \cite{land13122175, 10.3389/feart.2023.1133901}. The model is validated against these cooling effects to assess its ability to replicate observed processes. It is then used to forecast urban temperatures under future climate scenarios and to simulate targeted in-painting as a strategy to mitigate UHI impacts. The result is a unified workflow that integrates data analysis, forecasting, and simulation to inform climate-resilient urban planning.

\section{Methodology}
GFMs have demonstrated strong performance in geospatial tasks, with promising generalization across space and time, making them suitable for UHI applications in the context of increasing urbanization and climate-induced temperature extremes \cite{jakubik2023foundation}. However, existing studies rarely conduct in-depth evaluations of physical realism for GFMs, such as examining how urban green spaces influence modelled UHI dynamics through simulations like inpainting or future-scenario forecasting, which are essential for informed urban heat mitigation and planning \cite{fuPredictionSurfaceUrban2024}. To address this gap, this study follows a three-phase experimental workflow. 

First, an empirical baseline of how green spaces influence temperature is established by overlaying land-use/land-cover (LULC) maps with high-resolution land surface temperature (LST) imagery to extract park areas and analyse how urban temperature gradients evolve with distance from these green spaces, thereby quantifying internal cooling within parks and spillover cooling into surrounding built-up areas, with all results aggregated over summer daytime data from 2017 to 2025 across all cities, and assessing how well the GFM is able to replicate these effects for mitigation planning.\\
Second, the model’s extrapolation ability under future climate conditions is assessed through a focused experiment in Brașov, Romania, a city intentionally selected because it was not part of the original training set. Although GFMs are designed to generalize across regions, isolating a single unseen city allows for a clearer evaluation of performance and highlights the feasibility of future climate projections. In this setting, the data are ordered by temperature and the model is fine-tuned only on the cooler 90\% of observations, while the hottest 10\% are withheld to test generalization to extreme heat in Brașov. The difference between predicted and observed values in this upper percentile serves as an estimate of how far the model can reliably extrapolate beyond its training range. Under the assumption that this 10\% threshold represents its extrapolation limit, the model is then used to project future UHI conditions by replacing ERA5 with EURO-CORDEX climate inputs  under different future Representative Concentration Pathway (RCP) scenarios.\\
Third, a hypothetical urban greening intervention is simulated by guided inpainting of the satellite image, replacing built-up pixels with those representing urban green areas. The input spectral indices are therefore adjusted accordingly, and the model is run to predict the resulting thermal impact to support simulated UHI mitigation planning.

\subsection{Training procedure and geospatial foundation model}
Building on the work of \citet{10641750}, the Granite-Geospatial-Land-Surface-Temperature (Granite-GFM) model \cite{bhamjeeHuggingFaceGranite} is used as a GFM for predicting LST at high spatial and temporal resolution. The model is evaluated in two configurations, V1 which is fine-tuned using Harmonised Landsat Sentinel-2 (HLS L30 \cite{Masek_HLS_Sentinel2_2021}) imagery and ERA5-Land \cite{MunozSabater2019} 2 m air temperature statistics from 2013–2023 across 28 cities, and V2 which is fine-tuned on an extended dataset covering 52 cities across a wider range of hydro-climatic zones. Granite-GFM incorporates a Shifted Window (SWIN) Transformer architecture and builds on the Prithvi-SWIN-L Earth Observation foundation model \cite{Liu_2021_ICCV, Prithvi-100M,jakubik2023foundation}. While state of the art approaches for LST prediction focus on single regions of interest, the Granite-GFM model enables the estimation of LST at a 30m spatial and hourly temporal scale, for any city of interest using fewer input samples.

\subsection{Study regions and data}
Our analysis covers thirteen European cities that were not part of the training data, selected to test the model’s out-of-sample generalization across diverse urban environments. A regional map of the selected cities is provided in Appendix \ref{fig:appendix_study_regions_map}. Twelve of these cities, marked in red, are used in the first workflow step to assess internal and spillover cooling effects across varying hydro-climatic settings. For the second step, Brașov (marked in blue) is used as an unseen evaluation city. In the third intervention step, Prague is selected for in-painting. While these cities are highlighted, the workflow can in principle be applied to any urban area, which reflects the core advantage of geospatial foundation models. The methodology is based on four primary data sources. (1) Harmonised Landsat Sentinel-2 imagery providing 30\,m multispectral bands to derive LST using a split-window algorithm \cite{Masek_HLS_Sentinel2_2021, duPracticalSplitWindowAlgorithm2015}, (2) Impact Observatory 10\,m LULC data for identifying urban and green spaces \cite{karra2021global}, (3) ERA5-Land reanalysis providing continuous atmospheric context, with near-surface air temperature stacked onto the HLS imagery \cite{MunozSabater2019}, (4) EURO-CORDEX regional climate projections providing future forcing under multiple RCPs to represent different greenhouse-gas emission trajectories \cite{jacobEUROCORDEXNewHighresolution2014}.

\section{Experiments and results}
This section presents the experimental evaluation of our three-phase workflow with Model V1 and Model V2. The model’s predictive fidelity is assessed through the cooling anomaly ($\Delta T$), defined as the temperature difference between a point of interest and the surrounding built-up baseline. This isolates cooling effects from background variation and measures how well the model captures green space cooling. The difference between ground truth and predicted anomalies is quantified using Mean Absolute Error (MAE), Root Mean Square Error (RMSE), and Mean Bias Error (MBE). Forecasting performance is separately evaluated based on the model’s ability to extrapolate to extreme conditions, using MAE, Mean Square Error (MSE), and RMSE for the LST of the hottest 10\% of unseen data.

\subsection{Ground truth internal and spillover cooling patterns}
\begin{figure}[H]
    \centering
    \begin{subfigure}[b]{0.4\textwidth}
        \centering
        \includegraphics[width=\textwidth]{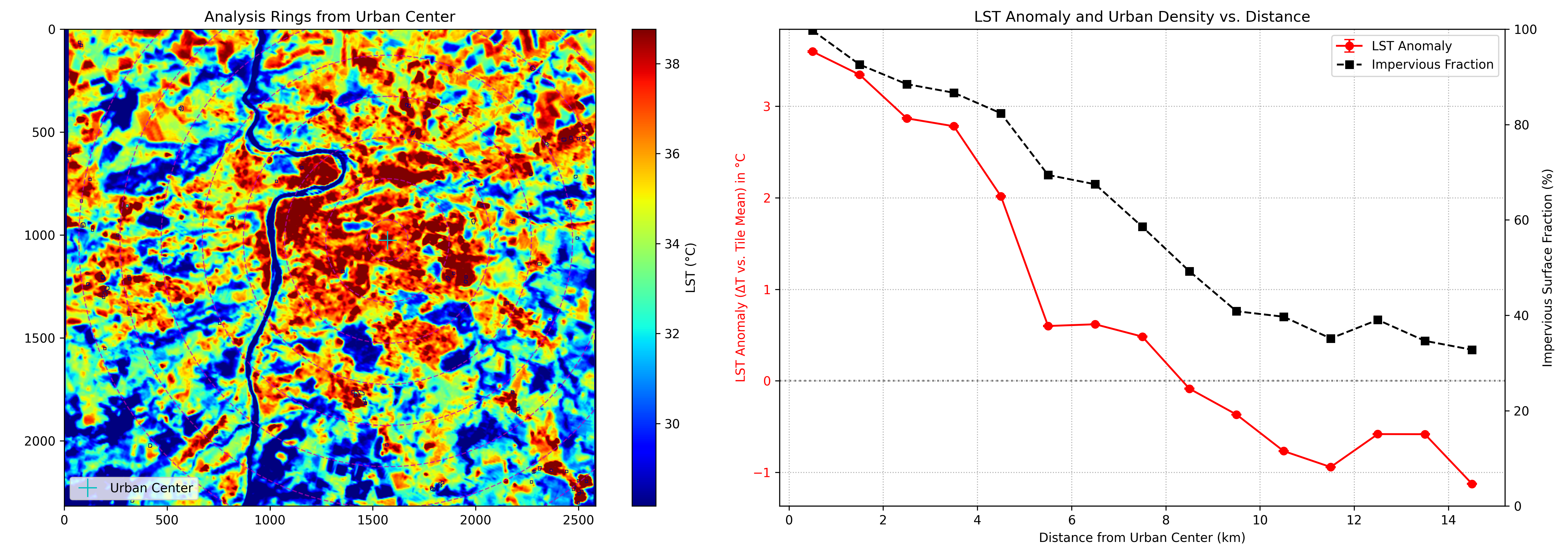}
        \caption{Urban gradient experiment}
        \label{fig:uhi_gradient}
    \end{subfigure}
    \hfill
    \begin{subfigure}[b]{0.4\textwidth}
        \centering
        \includegraphics[width=\textwidth]{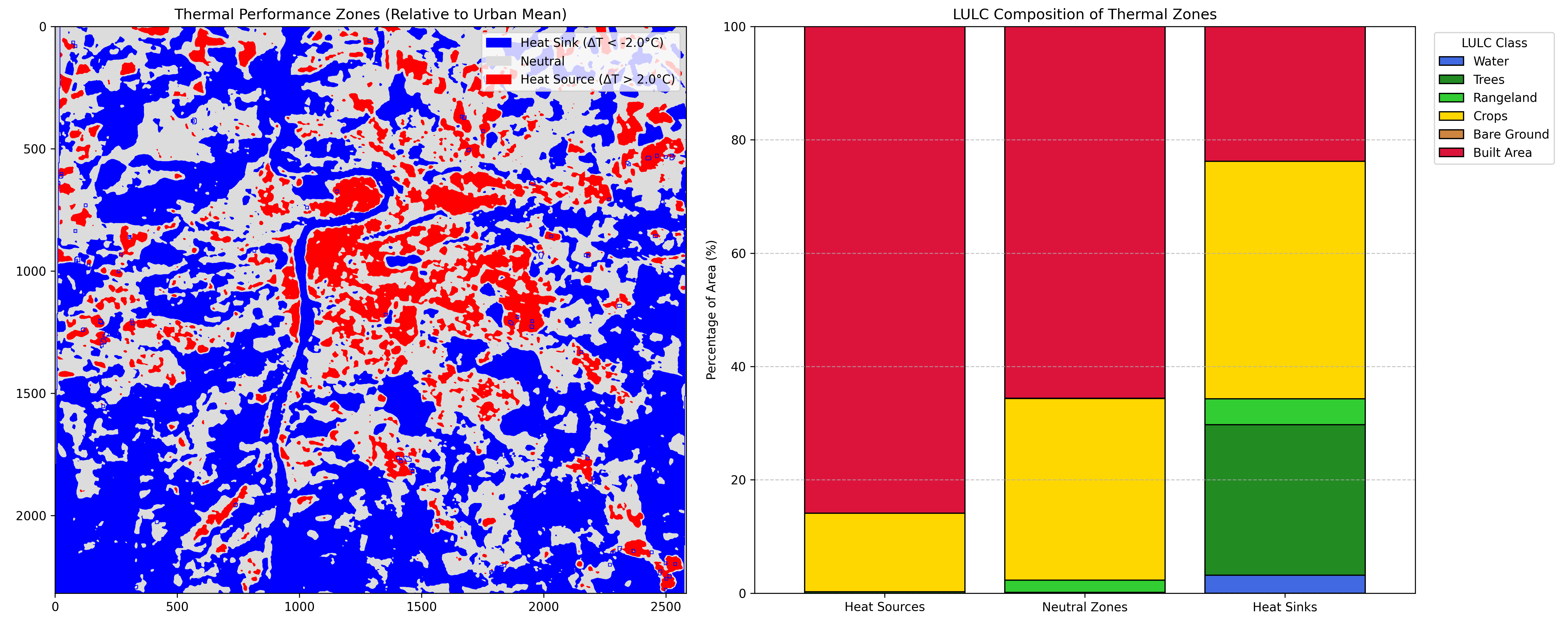}
        \caption{Heat source-sink experiment}
        \label{fig:source_sink}
    \end{subfigure}
    \caption{Baseline UHI characterization for Prague based on gradient and source–sink experiments.}
    \label{fig:uhi_gradient_sink}
\end{figure}

The ground truth analysis reveals clear thermal patterns shaped by urban green spaces. At the macro-scale, the urban gradient shows an average UHI of $+3.3^{\circ}\mathrm{C}$ in dense cores, driven by the lack of vegetation, as can be seen in Figure \ref{fig:uhi_gradient}. The heat source-sink analysis reinforces this link, with all tree cover located in the city’s coolest areas, as shown in Figure \ref{fig:source_sink}. At the micro-scale, parks exhibit strong internal cooling, with temperatures dropping up to $-2.6^{\circ}\mathrm{C}$ within 200 meters of the edge. Beyond their boundaries, spillover cooling lowers nearby temperatures by as much as $-3.5^{\circ}\mathrm{C}$ close to parks, decreasing to about $-1^{\circ}\mathrm{C}$ at 150 meters distance.

\subsection{Model performance evaluation and future climate forecasting}
Although both models achieve high accuracy in replicating the internal cooling gradient, model V2 exhibits a substantial improvement in representing the more complex spillover cooling phenomenon. Specifically, the MAE decreases from $0.302^{\circ}\mathrm{C}$ in V1 to $0.199^{\circ}\mathrm{C}$ in V2. As shown in Figure \ref{fig:cooling_gradients} and summarized in Table \ref{tab:key_exp_variants} in the Appendix, this demonstrates a statistically significant enhancement in performance attributable to the more diverse training of model V2.

\begin{figure}[h!]
    \centering
    \begin{subfigure}[b]{0.32\textwidth}
        \centering
        \includegraphics[width=\textwidth]{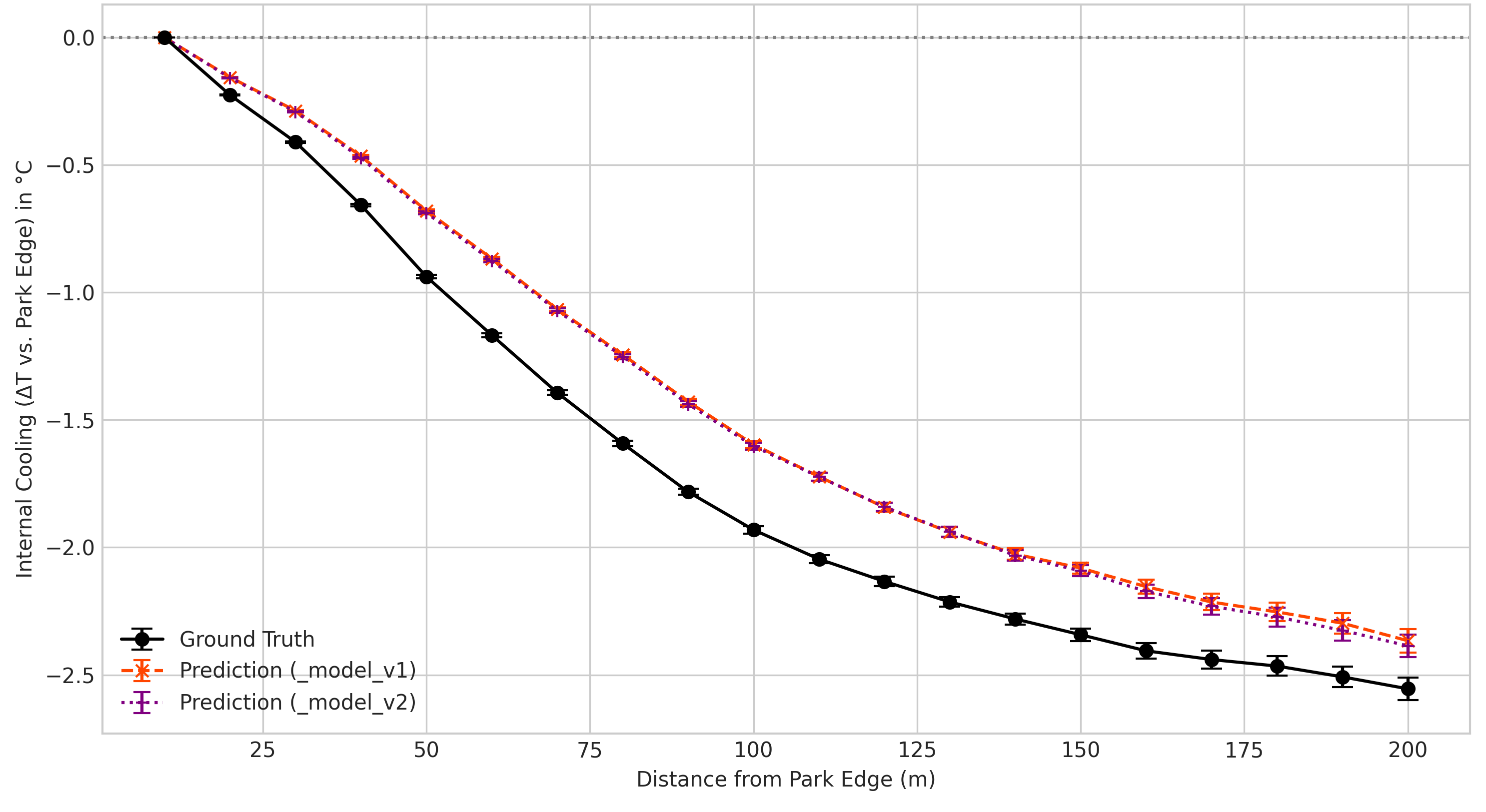}
        \caption{Internal cooling gradient}
        \label{fig:internal_gradient}
    \end{subfigure}
    \hfill
    \begin{subfigure}[b]{0.32\textwidth}
        \centering
        \includegraphics[width=\textwidth]{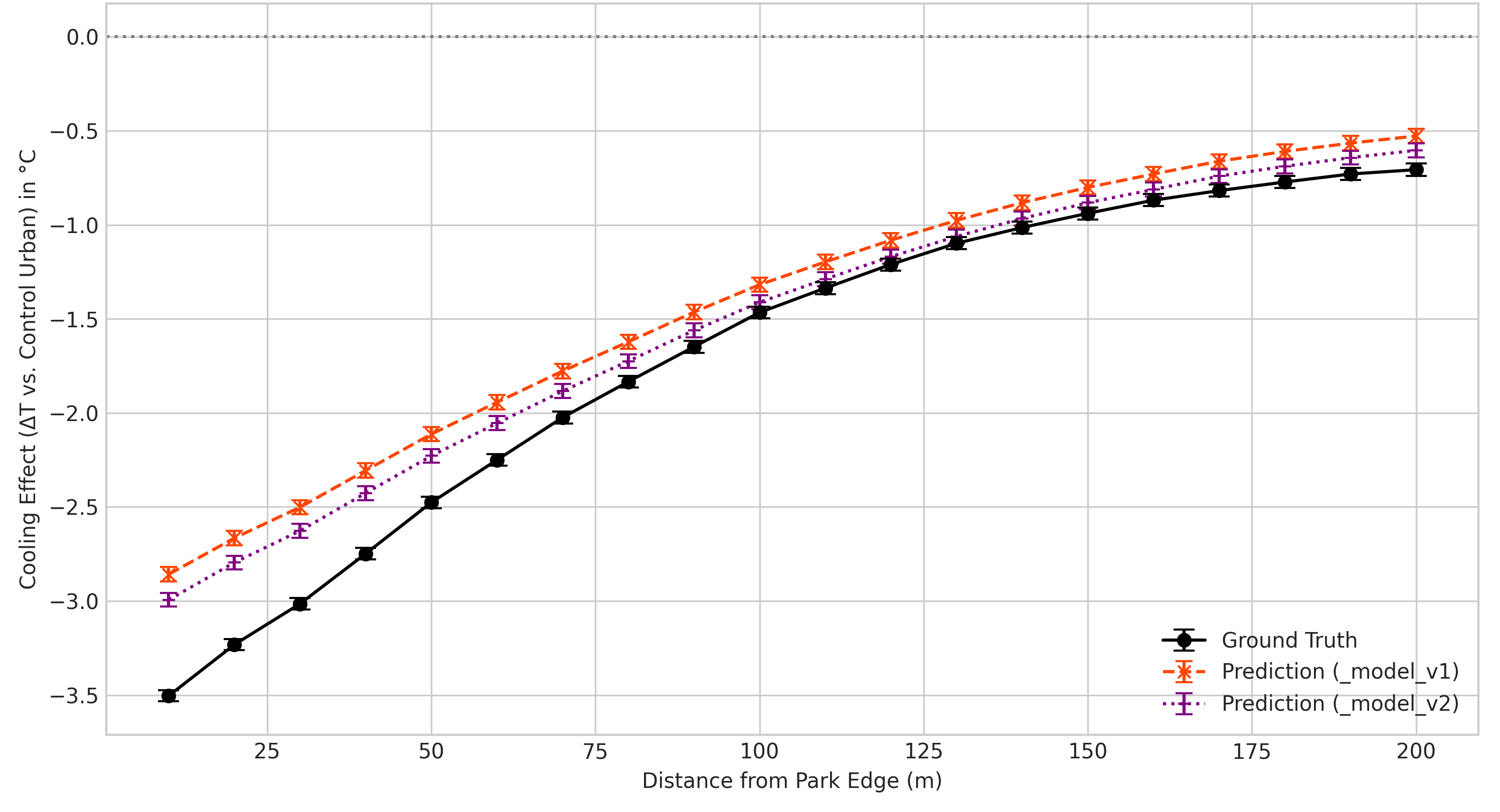}
        \caption{Spillover cooling gradient}
        \label{fig:spillover_gradient}
    \end{subfigure}
    \caption{Comparison of ground truth cooling phenomena with model predictions.}
    \label{fig:cooling_gradients}
\end{figure}

\begin{figure}[h!]
    \centering
    \includegraphics[width=0.33\textwidth]{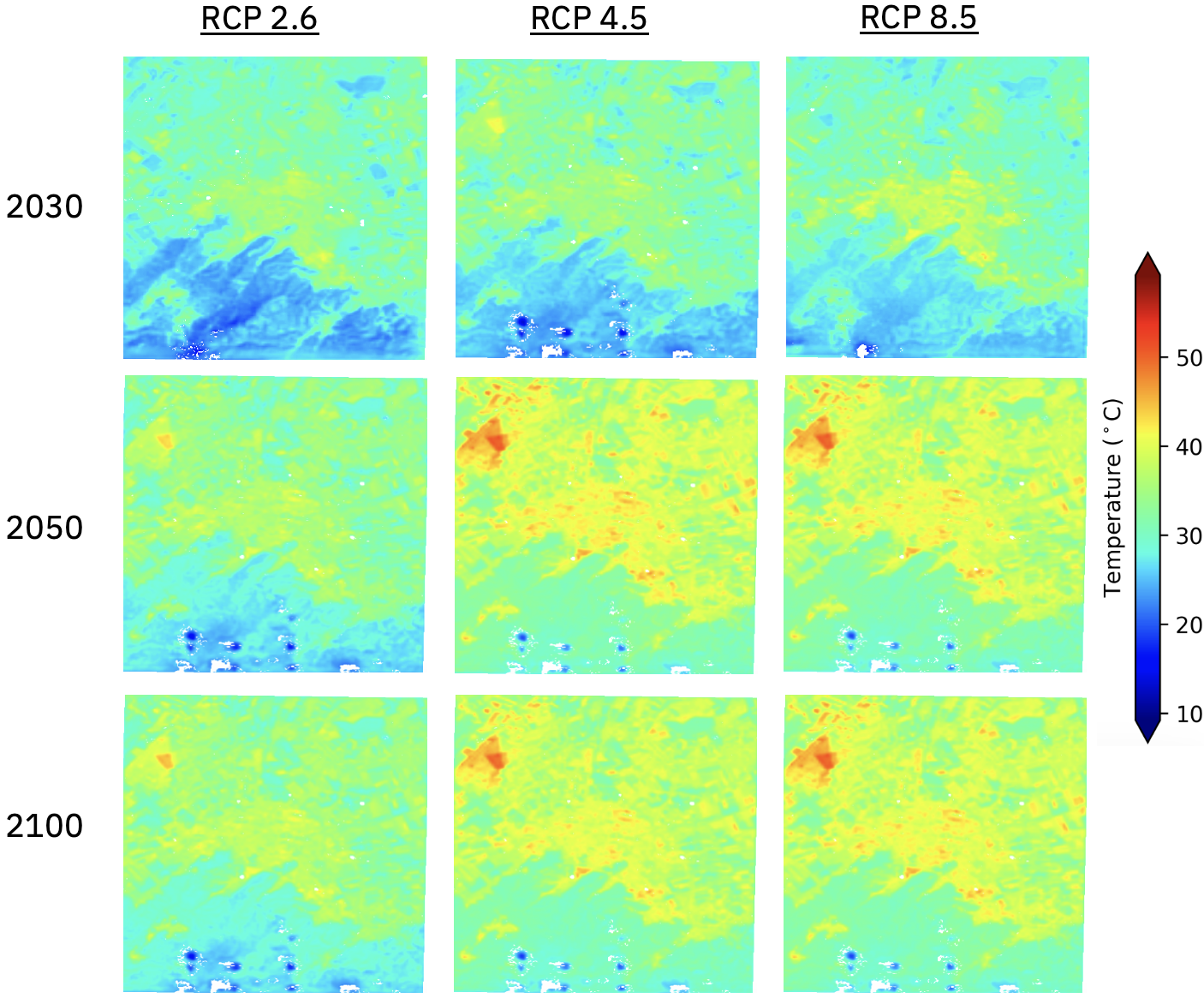}
    \caption{Projected UHI extent under RCP 2.6, 4.5, and 8.5 for 2030, 2050, and 2100 in Brașov.}
    \label{fig:rcp_scenarios}
\end{figure}

Furthermore, the model shows strong extrapolation performance, achieving an MAE of 1.74 °C on the upper 10\% of unseen temperature data using model V2. This improved accuracy, particularly in extrapolating to extreme conditions, provides strong confidence in its predictive capabilities for forecasting and simulation tasks. The climate projections in Figure \ref{fig:rcp_scenarios} indicate a substantial intensification of urban heat islands under RCP 4.5 and 8.5 and show the development of UHI in Brașov for 2030, 2050, and 2100 under RCP 2.6, 4.5, and 8.5, based on EURO-CORDEX data.

\subsection{Urban greening simulation through inpainting}
\begin{figure}[h!]
    \centering
    \includegraphics[width=0.43\textwidth]{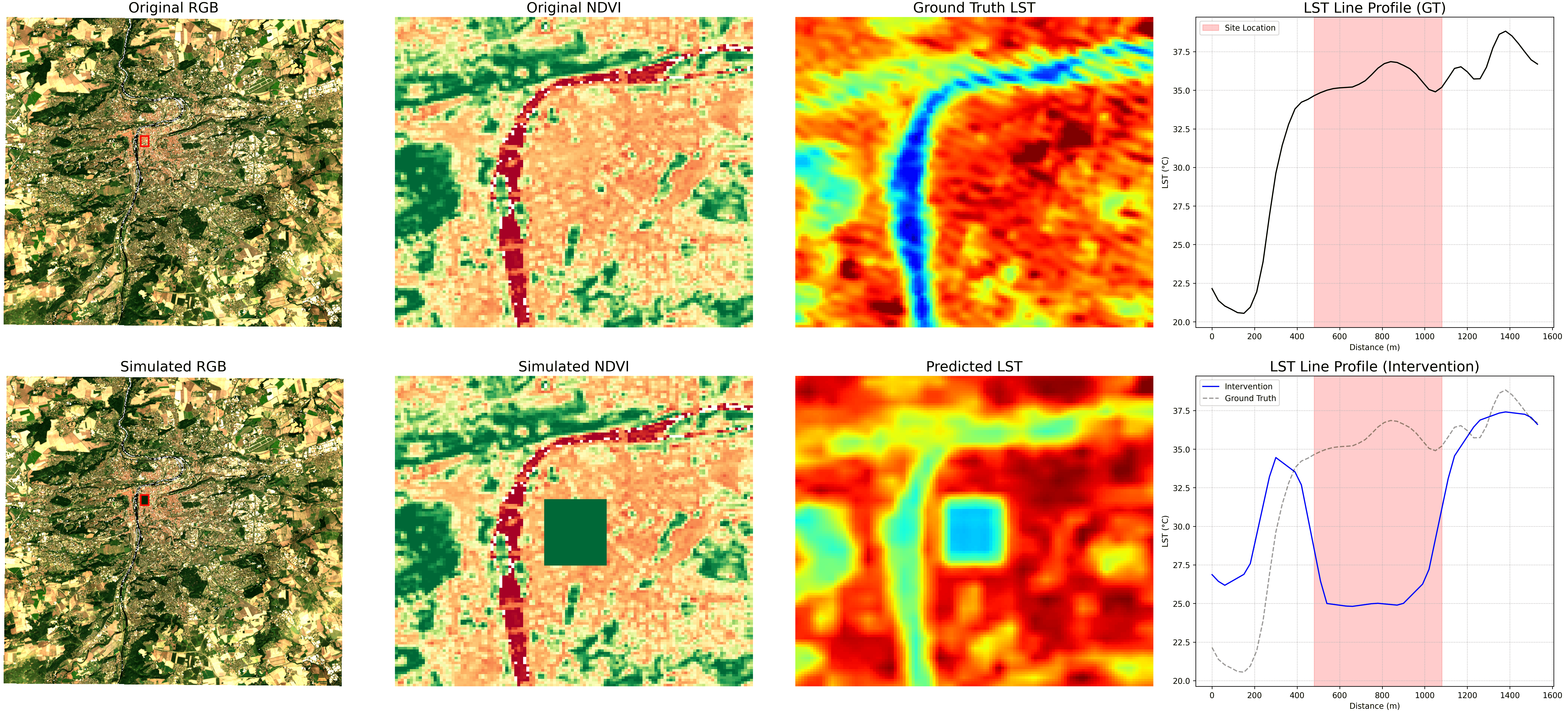}
    \caption{Comparison of inpainting results against ground truth data in Prague.}
    \label{fig:inpainting_results}
\end{figure}

After establishing the model’s capacity to replicate real-world cooling effects and forecast scenarios with high accuracy, the effectiveness of Model V2 is demonstrated through an urban greening scenario. The simulation provides a “before and after” comparison of the intervention area across RGB-, normalized difference vegetation index (NDVI)-, and LST imagery, as can be seen in Figure \ref{fig:inpainting_results}, together with a transect showing the LST profile through the intervention area, confirming the land cover change and its cooling effect. Building on this, the simulation further quantifies the internal cooling within the new park and the spillover into the surrounding neighborhood, as shown in Figure \ref{fig:inpainting_results_2} in the Appendix, demonstrating the model’s ability to provide evidence-based metrics for urban planning decisions.

\section{Conclusion}
This paper demonstrates a comprehensive workflow using a fine-tuned GFM to analyze, forecast, and simulate mitigation strategies for UHI. We have shown that by first establishing an empirical ground truth, we can effectively validate a GFM's ability to replicate complex thermal phenomena. The validated model can then be applied to forecast future climate scenarios and to perform in-painting for assessing the impact of potential interventions. This approach transforms the GFM from an assessment tool into an interactive simulation platform for building climate-resilient cities.

\begin{ack}
This work was supported by funding from the European Union's Horizon 2020 research and innovation programme for the project NATURE-DEMO under grant agreement No. 101157448, and by the Swiss State Secretariat for Education, Research and Innovation (SERI) contract 24.00279 (REF-1131-52105).
\end{ack}

\small
\bibliographystyle{unsrtnat}
\bibliography{references}
\normalsize
\newpage

\appendix
\section{Appendix}

\subsection{Overview of selected study regions}

For the experiments, twelve European cities were selected as study regions. For each city, Harmonized Landsat Sentinel-2 (HLS) imagery, ERA5-Land atmospheric reanalysis data, ESRI land use/land cover classifications, and EURO-CORDEX climate scenario data were downloaded and processed, covering the historical period from 2017 to 2025 and future projections for 2030, 2050, and 2100.
Together, these datasets ensure consistent coverage, high resolution, and comparability across cities, supporting both ground-truth analysis and predictive modeling.

\begin{figure}[H]
    \centering
    \includegraphics[width=0.8\textwidth]{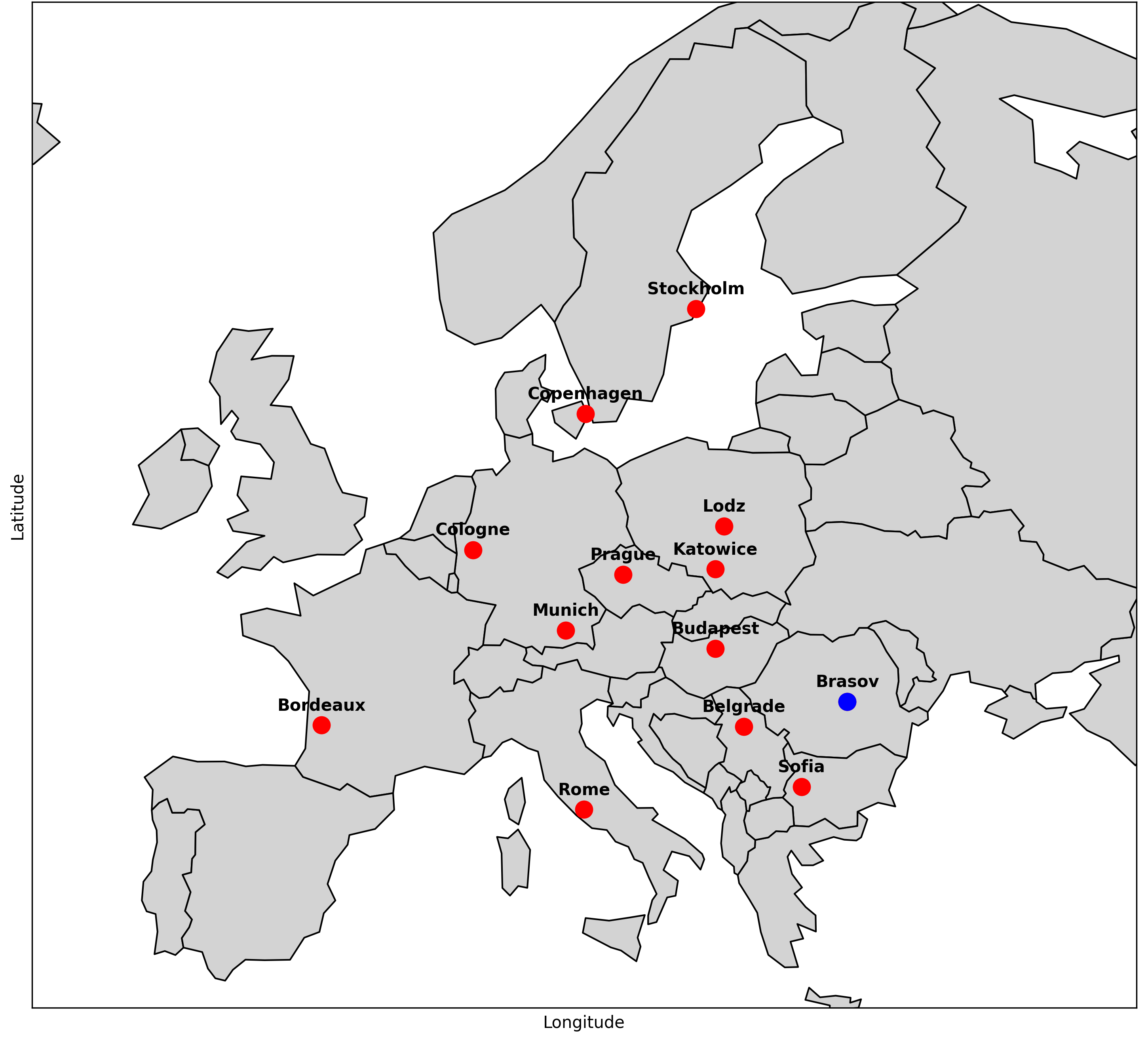}
    \caption{Map of the twelve European study cities (red dots), selected for their diverse urban characteristics, hydro-climatic settings, and suitability within the HLS tiling system, which were further analyzed for their internal and spillover cooling effects. Brașov (blue dot) was chosen as an extrapolation site to test model generalization.}
    \label{fig:appendix_study_regions_map}
\end{figure}

\subsection{Evaluation of urban green area cooling effects using a geospatial foundation model}

The predictive framework’s ability to capture urban cooling effects was assessed by benchmarking the two model variants, V1 and V2, based on their performance in representing internal cooling gradients within parks and the spillover of cooling into surrounding urban areas. Model accuracy was assessed using MAE, RMSE, and MBE, enabling direct comparison of predictive performance across both experiments.

\begin{table}[H]
\centering
\caption{Performance comparison of model variants for the two key cooling experiments. The best value per metric between V1 and V2 is \underline{underlined}.}
\label{tab:key_exp_variants}
\begin{tabular}{@{}l c c c c c c@{}}
\toprule
 & \multicolumn{3}{c}{\textbf{V1}} & \multicolumn{3}{c}{\textbf{V2}} \\
\cmidrule(lr){2-4} \cmidrule(lr){5-7}
\textbf{Experiment} & MAE & RMSE & MBE & MAE & RMSE & MBE \\
\midrule
Internal Cooling          & 0.240 & 0.257 & +0.240 & \underline{0.231} & \underline{0.249} & \underline{+0.231} \\
Spillover Cooling         & 0.302 & 0.339 & +0.302 & \underline{0.199} & \underline{0.243} & \underline{+0.199} \\
\bottomrule
\end{tabular}
\end{table}

\subsection{Analysis of cooling gradients for urban green areas using a geospatial foundation model}

The thermal impact of the simulated greening intervention was assessed using Model V2 by analyzing both internal and external cooling responses. The internal cooling gradient captures the temperature decrease from the park edge toward its center, while the spillover gradient reflects how far this effect extends into surrounding built-up areas. Together, these analyses demonstrate both local and neighborhood-scale cooling benefits.

\begin{figure}[H]
    \centering
    \begin{subfigure}[b]{0.49\textwidth}
        \centering
        \includegraphics[width=\textwidth]{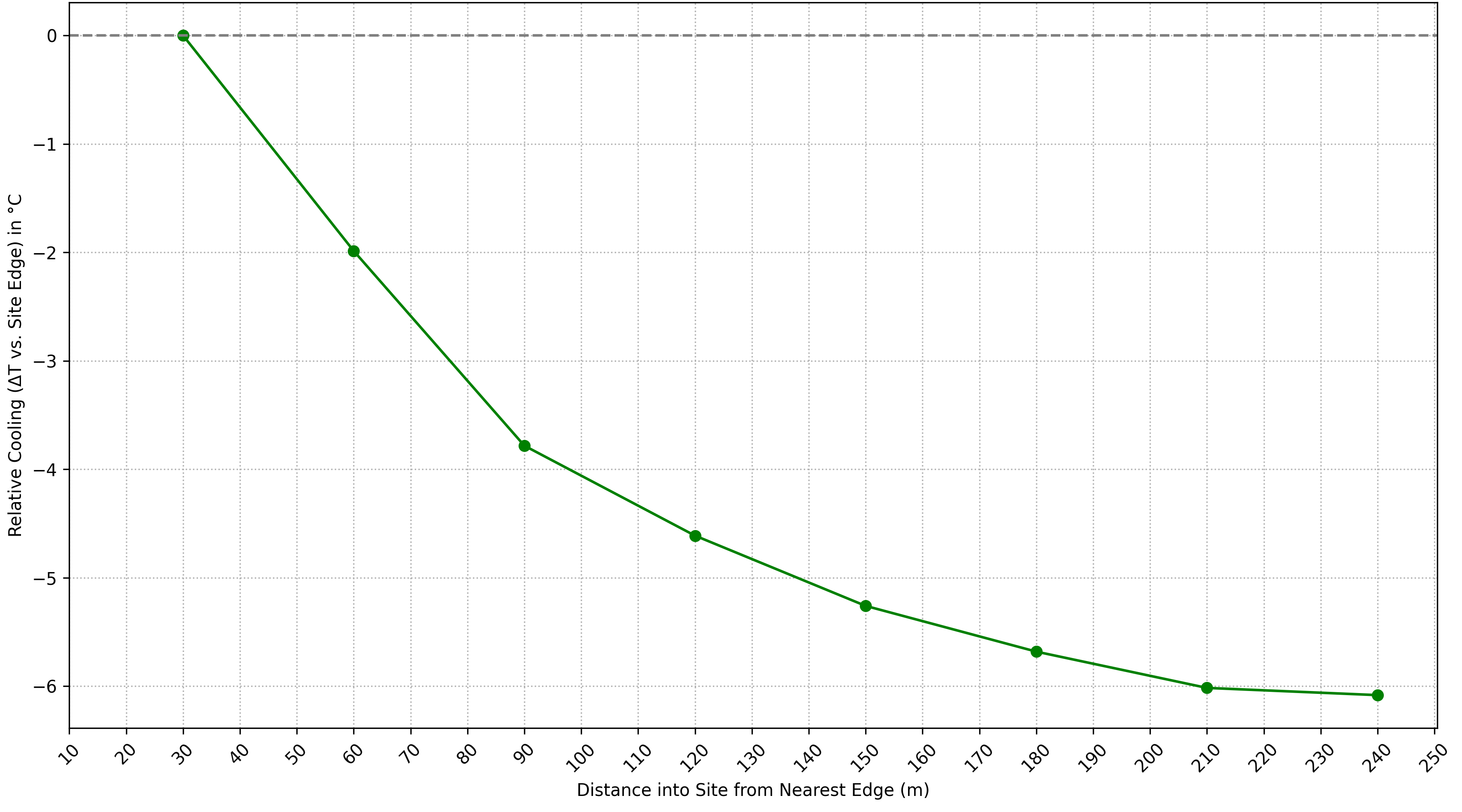}
        \caption{Internal Cooling Gradient}
        \label{fig:internal_gradient_intervention}
    \end{subfigure}
    \hfill
    \begin{subfigure}[b]{0.49\textwidth}
        \centering
        \includegraphics[width=\textwidth]{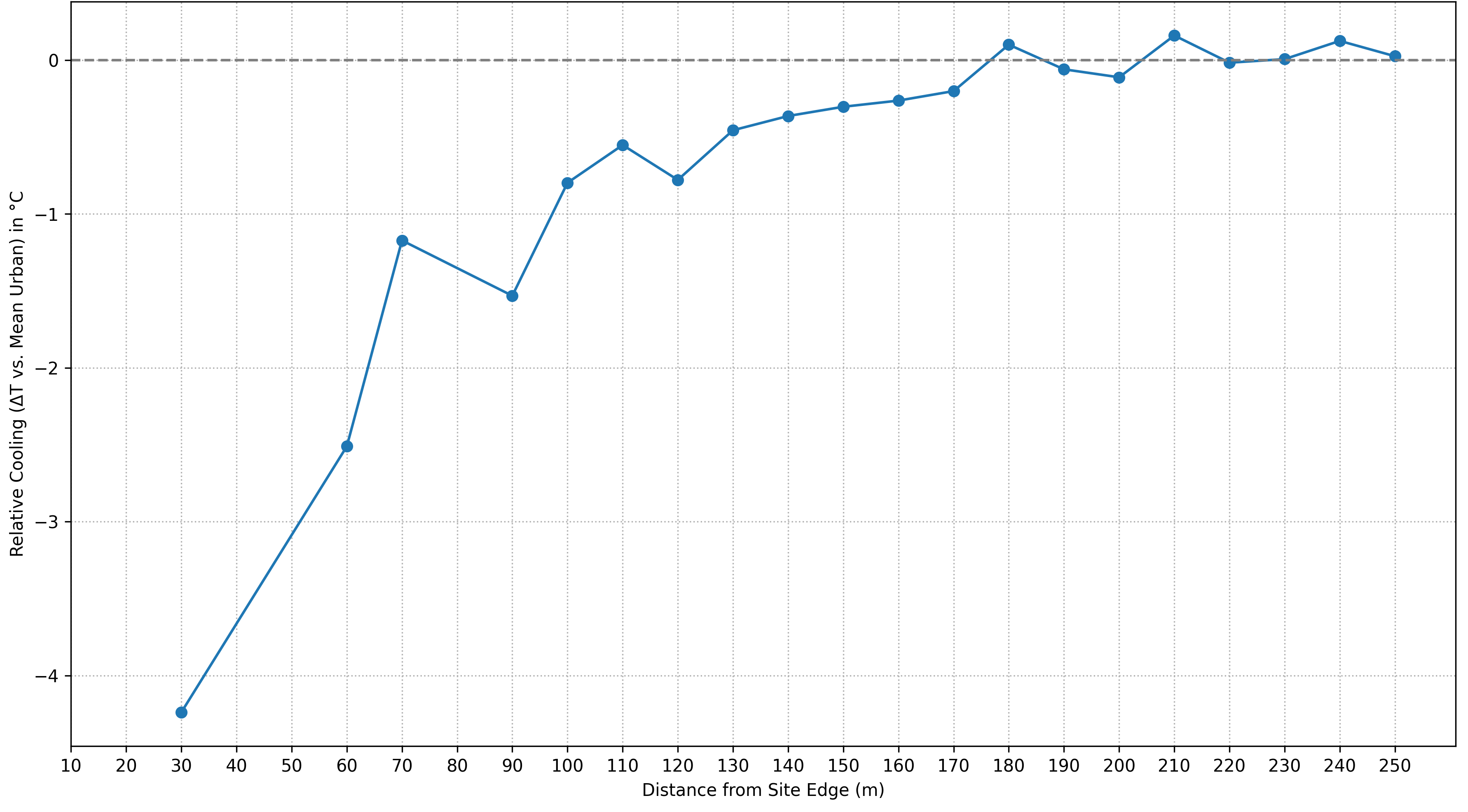}
        \caption{Spillover Cooling Gradient}
        \label{fig:spillover_gradient_intervention}
    \end{subfigure}
    \caption{Cooling phenomena of the proposed intervention area.}
    \label{fig:inpainting_results_2}
\end{figure}

\subsection{Comparison of model variants for extrapolation capability}

Table~\ref{tab:model_comparison} presents the test set results for Model V1 and Model V2, both fine-tuned on Brașov-specific data using identical hyperparameters. Three evaluation setups were applied. The \textit{Baseline} reflects zero-shot performance without Brașov-specific fine-tuning. The \textit{Random Data Split} follows a conventional 72/18/10 train/validation/test split. The \textit{High Heat Scenario} implements an extrapolation experiment by restricting training and validation to the lower 90\% of temperature observations and reserving the upper 10\% for testing, thereby assessing how well the models predict land surface temperatures beyond their training range.

\begin{table}[H]
  \centering
  \caption[Model comparison via MAE, MSE, RMSE]{Performance comparison of model variants using MAE, MSE, and RMSE. Bold indicates the best value and underline indicates the second-best.}
  \begin{tabular}{@{}l ccc ccc@{}}
    \toprule
     & \multicolumn{3}{c}{\textbf{V1}} & \multicolumn{3}{c}{\textbf{V2}} \\
    \cmidrule(lr){2-4} \cmidrule(lr){5-7}
    \textbf{Model Variants} & MAE & MSE & RMSE & MAE & MSE & RMSE \\
    \midrule
    Baseline                 & 1.95 & 7.25 & 2.69 & 2.81 & 12.94 & 3.60 \\
    Random Data Split        & 1.80 & \underline{6.26} & \underline{2.50} & \underline{1.77} & \textbf{5.80} & \textbf{2.41} \\
    High-Heat Scenario (90th) & 1.96 & 7.05 & 2.66 & \textbf{1.74} & 6.37 & 2.52 \\
    \bottomrule
  \end{tabular}
  \label{tab:model_comparison}
\end{table}

The results indicate that the \textit{High Heat Scenario} models perform comparably to the random split across all metrics, while consistently exceeding the \textit{Baseline}, except for MAE in V1. Notably, V2 achieves lower MAE in the \textit{High Heat Scenario} than in the random split, demonstrating a stronger capacity for generalization. Across all settings and model variants, differences remain below 0.60\,${}^\circ$C, showing that GFMs can extrapolate beyond training data and capture high-heat extremes relevant for UHI dynamics.

For Brașov, the 90th percentile temperature was 23.29\,${}^\circ$C, with the GFM extrapolating up to 26.91\,${}^\circ$C and reaching an MAE of 1.74\,${}^\circ$C. This represents a successful prediction of 3.62\,${}^\circ$C beyond the training limit, which provides an estimate of the model’s extrapolation limit, forming the basis for projecting future UHI intensity under EURO-CORDEX climate scenarios.

\end{document}